\documentclass[runningheads]{llncs}
\usepackage{graphicx}
\usepackage{amsmath}
\usepackage{amssymb}
\usepackage{multirow}
\usepackage{CJKutf8}

\usepackage{enumitem}
\begin{document}
\title{Distant finetuning with discourse relations for stance classification}
%
%
\author{Lifeng Jin \and
Kun Xu \and
Linfeng Song \and
Dong Yu}
\authorrunning{L. Jin et al.}
%
\institute{Tencent AI Lab, Bellevue WA 98003, USA
\email{\{lifengjin,kxkunxu,lfsong,dyu\}@tencent.com}}
\maketitle              
\begin{abstract}
Approaches for the stance classification task, an important task for understanding argumentation in debates and detecting fake news, have been relying on models which deal with individual debate topics. In this paper, in order to train a system independent from topics, we propose a new method to extract data with silver labels from raw text to finetune a model for stance classification. The extraction relies on specific discourse relation information, which is shown as a reliable and accurate source for providing stance information. We also propose a 3-stage training framework where the noisy level in the data used for finetuning decreases over different stages going from the most noisy to the least noisy. Detailed experiments show that the automatically annotated dataset as well as the 3-stage training help improve model performance in stance classification. Our approach ranks 1$^{\text{st}}$ among 26 competing teams in the stance classification track of the NLPCC 2021 shared task Argumentative Text Understanding for AI Debater, which confirms the effectiveness of our approach.

\end{abstract}
\begin{CJK*}{UTF8}{gbsn}

\section{Introduction}

In natural language understanding, it is important to understand how sentences are used in order to argue for or against particular topics in conversations and articles. This not only relates to automatic debates \cite{Slonim2021-bg} but also is useful in detecting fake news in media and allowing colorful persona in robots \cite{Kucuk2020-fa}. Stance detection or stance classification \cite{Bar-Haim2017-hv,Kucuk2020-fa} is the task where one has to decide whether a given claim is in support of or against a given topic, or the two are unrelated in terms of argumentation. The support and against relations between topics and claims are usually more abstract than such relations in opinion mining, because in instead of directly taking or refuting a topic, a claim is usually a piece of evidence or a logical consequence following a stance towards some topic, which makes detecting the stance of such claims difficult and knowledge-intensive. This problem is partially tackled by approaches where topic-specific models are used. Obviously it is difficult to generalize to new topics with these models, because new models have to be trained with annotated data for the new topics, and possible topics in real life scenarios are numerous. Generalizability is also a problem for machine learning models from the stand point of training data, because common stance detection datasets have only a couple hundred topics but thousands of claims, allowing such models to easily overfit to the topics in training data.

We propose to address the generalizability issue as well as the knowledge-intensive nature of the task with knowledge-rich pretrained models. Pretrained models have shown good performance in a variety of natural language understanding tasks which require both linguistic and commonsense knowledge. Such knowledge is invaluable to the stance detection task. In order to further improve model performance, we extract a noisy training dataset from large quantities of unlabeled text, following the intuition that discourse relations are indicative of stance in general. For example, the relationship between a topic, such as ``大数据带来了更多的好处 (big data brings more good than bad)'', and a supporting argument, such as ``生成的大数据可作为预测工具和预防策略 (the generated big data can be used as a predictive tool and preventive strategy)'', may be rewritten as a causal relation: 
\begin{enumerate}
    \item 因为生成的大数据可作为预测工具和预防策略，所以大数据带来了更多的好处。(Because the generated big data can be used as a predictive tool and preventive strategy, big data brings more good than bad.),
\end{enumerate}
and the same topic and an against argument, such as ``大数据的准确性难以确保 (the accuracy of big data is hard to be sure of)'', may be rewritten as a contradiction relation: 
\begin{enumerate}[resume]
    \item 虽然大数据带来了更多的好处，大数据的准确性难以确保。(Although big data brings more good than bad, the accuracy of big data is hard to be sure of.),
\end{enumerate}
which suggests that raw sentences in such relations may be in turn used as noisy training instances for the stance detection task.

Training neural network models with such noisy datasets improves robustness of the model, reduces greatly the chance of overfitting, allows the model to acclimate to task-specific data format, and provides chances to learn more knowledge for the stance detection task. Experiments on development data show large improvements over baselines where such noisy data is not used. Amongst the 26 teams participating in the Claim Stance Classification for Debating track of the Argumentative Text Understanding for AI Debater shared task, our approach ranks 1$\text{st}$, with 2.3\% absolute performance improvement over the runner-up approach.

\section{Related work}

Stance classification has been a subject of research in many different environments, such as congressional debates \cite{Thomas2006-dc}, online debates on social media \cite{Somasundaran2009-id,Conforti2020-kc} and company-internal discussions \cite{Murakami2010-lf}. Previous approaches focus on learning topic-specific models to classify stances of related claims with machine learning models \cite{Anand2011-qu,Hasan2013-ek,Sobhani2016-gd} as well as deep learning models \cite{Sun2018-ea,Dey2018-xa,Ghosh2019-ou,Popat2019-nf,Sirrianni2020-lt,Yu2020-mo}. Previous work has also looked at doing stance classification at challenging situations such as zero-shot \cite{Allaway2020-ua} and unsupervised settings \cite{Somasundaran2009-id,Ghosh2018-rx,Kobbe2020-nq}. Since stance classification has been thought of as a subtask of sentiment analysis \cite{Kucuk2020-fa}, the use of sentiment lexicon is popular in previous work.
Compared to previous work, our approach does not rely on any sentiment lexicon, which is a linguistic resource difficult to construct. Our approach also does not require topic-specific model training, which improves generalizability of a trained model to unseen topics and claims.

\section{Unsupervised Data Preparation}

We follow the intuition that the Support relation in stance classification between claims and topics can be categorized as a causal or conditional relation, because one should be able to deduce the topic from the claim if the claim supports the topic. Similarly, the Against relation between claims and topics can be categorized as a contraction relation where the claim does not naturally follow a topic. If a claim and a topic are to be connected by discourse connectives, connectives of corresponding discourse relations need to be used in order to preserve discourse coherence. Sentences with such discourse relations could better prepare the pretrained language models for finetuning with gold data and help the language models fight against overfitting. We first present a few different sets of data we extract from raw text with no supervision, and then explain how they are used in our finetuning framework.

\subsection{Data $D_1$ Extraction for Distant Finetuning}
\label{sec:data1_extraction}
A dataset for unsupervised distant finetuning is extracted from a large text corpus CLUE \cite{xu-etal-2020-clue}\footnote{\url{https://github.com/CLUEbenchmark/CLUE}} based on discourse relations. Table \ref{tab:connectives} shows examples of discourse connectives used for extracting sentences with particular discourse relations. A pair of sentences are kept when the second sentence starts with a multiple line connective, which follows this pattern ``$S_1\text{。} c_1 S_2\text{。}$'' where $S_i$ is a sentence or a sentence fragment, and $c_i$ is a discourse connective. For Support, $S_1$ is a topic and $S_2$ is a claim, where the opposite is adopted for Against.
For single sentences, one sentence is kept if it contains a pair of single line connectives where the second connective is in a sentence fragment directly after a comma, which follows this pattern ``$S_1 c_1 S_2\text{，} S_3 c_2 S_4\text{。}$''. In the case of single sentences, for Support, $S_1c_1S_2$ is a topic and $S_3c_2S_4$ is a claim, where the opposite is adopted for Against. Candidate sentences are discarded when they contain non-Chinese characters, exceed 100 characters, or contain pronouns. The discourse connectives are deleted from the sentences to remove obvious and easy cues to the relation classes. The sentence pairs with the Neutral label are selected randomly from sentences in the same article which are close to the topic sentence. The final $D_1$ dataset includes 1.2 million data points labeled as Support, 0.7 million labeled as Against, and 1.9 million labeled as Neutral. Table \ref{tab:example_sentences} shows examples of extracted sentences with different silver labels.

\begin{table}[]
\caption{Example Chinese discourse connectives used in extraction.}\label{tab:connectives}
    \centering
    \begin{tabular}{|l|l|l|}
    \hline
     Relation & Type  &  Connectives \\
     \hline
       \multirow{2}{*}{Support}  &  Multiple line & 因此, 因而, 所以\\
       & Single line & 因为...所以..., 只要...就..., 要是...就..., 之所以...是因为... \\
     \hline
       \multirow{2}{*}{Against}  &  Multiple line & 但是, 然而, 可是 \\
       & Single line & 虽然...但是..., 虽然...可是..., 尽管...但是... \\
     \hline
    \end{tabular}
\end{table}

\begin{table}[]
\caption{Examples of extracted sentence pairs from raw text.}\label{tab:example_sentences}
    \centering
    \begin{tabular}{|l|l|l|}
    \hline
     Relation & Type  &  Connectives \\
     \hline
       \multirow{2}{*}{Support}  &  Topic & 常将弹性工时与变形工时相互混淆\\
       & Claim & 国内学界对于弹性工时概念未有统一解释 \\
     \hline
       \multirow{2}{*}{Against}  &  Topic & 选择不同作用机制的癫痫药物，才可能获得疗效的叠加 \\
       & Claim & 如果两种癫痫药物有相同的不良反应，就不能联合使用 \\
    \hline
       \multirow{2}{*}{Neutral}  &  Topic & 其中的人数是最基本的数据 \\
       & Claim & 人口数据是一个国家和地区的基本数据 \\
     \hline
    \end{tabular}
\end{table}

\subsection{Low-noise Finetuning Data $D_2$ Extraction}
\label{sec:data2_extraction}

Although the distant finetuning data prepared in Section \ref{sec:data1_extraction} can provide training signal to further pretrain language models, it may be too noisy for final finetuning purposes. The Conditional relation does not always equal to Support, as portrayed in this example ``只要小明去，小张就会去。(If Xiao Ming goes, Xiao Zhang goes too.)'' in which the condition has only an arbitrary connection to the result. Similarly the Contradiction relation is not always Against, shown in this example ``虽然兔毛可以抵御严寒，但是兔子也怕热。(Although rabbit fur can be good for rigid cold, rabbits are also prone to overheating)'' where the two facts are more supplementary than contradictory to each other. Further filtering is needed to reduce the noise level within the extracted pretrain dataset.

A list of high frequency topic indicators is used to find sentences that are most likely to be statements of positions on certain issues, which are the best candidates for topcis. The list includes words such as ``应 (should)'' and ``最 (most)''. More importantly, we consider Entailment and Contradiction relations from the natural language inference (NLI) task very close to the Support and Against relations in stance detection, therefore we employ an NLI model for data selection. Specifically, a Chinese BERT with a classification layer is finetuned with the XNLI dataset on all available languages and the best model is chosen based on evaluation on the Chinese NLI portion of the XNLI evaluation dataset. This model is then used to make predictions of NLI labels on all data points in $D_1$. Finally, 30,000 data points which are either labeled Support by the connectives and Entailment by the XNLI model, or Against and Contradiction, or Neutral and No Entailment are randomly sampled from $D_1$, resulting in a low-noise finetuning dataset $D_2$ with 30,000 data points in total, which is about 5 times the size of the gold training set.

\subsection{Stance Detection Data in other languages}

Datasets for stance detection also exist in other languages such as English. With a pretrained language model able to take multilingual input, we expect such datasets help the model learn the concept of Support and Against more robustly. The multilingual stance detection dataset XArgMining \cite{Toledo-Ronen2020-tm} from the IBM Debater project contains human-authored data points for stance detection in English, as well as such data points translated into 5 other languages: Spanish, French, Italian, German and Dutch. With both human authored and machine translated data points combined, the dataset used for training has 400,000 data points. The dataset $D_x$ is the concatenation of these datasets.

\section{Staged Training with Noisy Finetuning}

Our model used for the task follows the standard pretraining-finetuning paradigm. A base transformer-based language model pretrained on large quantities of unlabeled data is used as an encoder to encode the topic and the claim. The contextualized embedding of the [CLS] token is used for classification, which goes through a linear layer to generate the logits for the three labels.

\begin{figure}
\includegraphics[width=\textwidth]{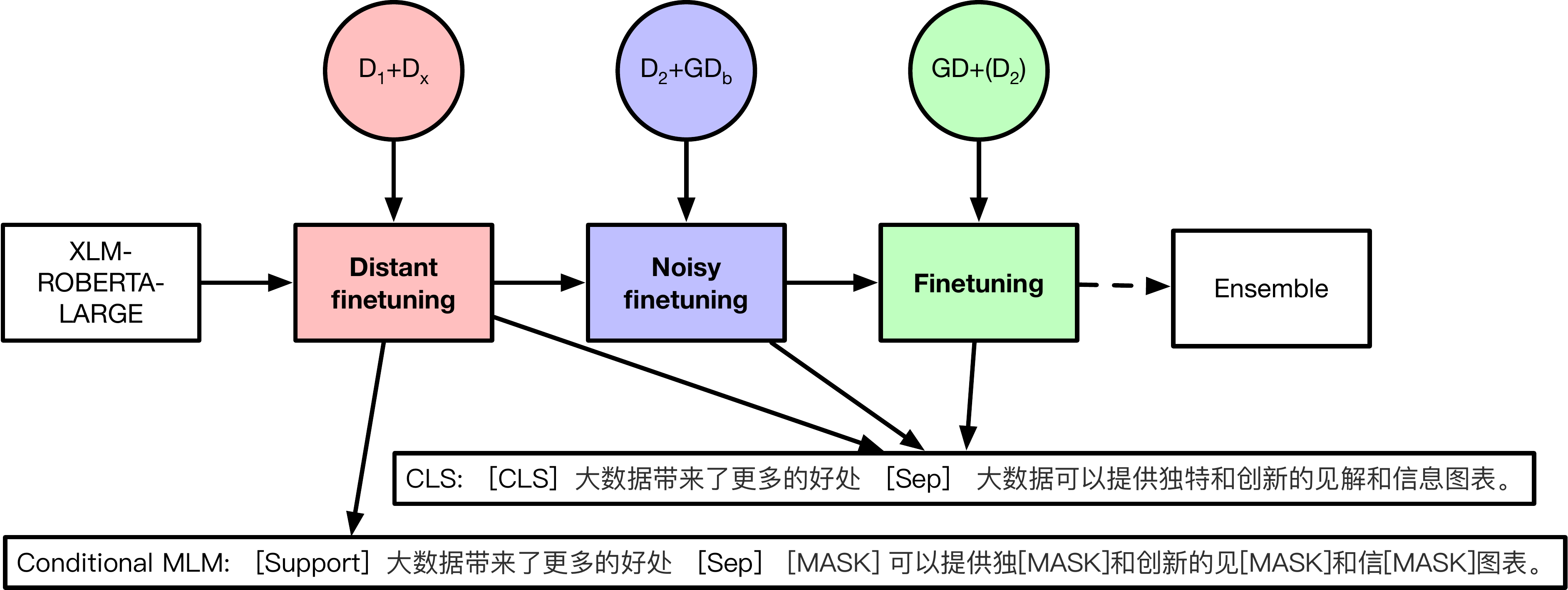}
\caption{The training process with noisy datasets.} \label{fig:train}
\end{figure}

In order to utilize the large amount of noisy data to help our model get better results, a novel training process where datasets with different noise levels are used in different stages to finetune the model, which is shown in Fig.~\ref{fig:train}. There are three stages in the whole finetuning process. The first stage is to use $D_1$ and $D_x$ for distant finetuning, and the second stage is to use the low-noise refined dataset and the back-translated gold dataset for noisy finetuning, and the final stage is to use the gold data with a small portion of noisy data for final finetuning.

\subsection{Distant finetuning}
\label{sec:distant_finetuning}
In this finetuning stage, datasets with high noise level $D_1$ or with data points in other languages $D_x$ are used as training data. There are two training objectives used in this stage: conditional masked language modeling and classification. For each batch of training data points, one training objective is randomly chosen. For the conditional masked language modeling objective, the topic sentence and the claim sentence are first concatenated and tokenized by a tokenizer from a pretrained language model, and then the [CLS] token at the beginning of the tokenized sequence is replaced by a special token indicating the label of the pair. 
Part of either the topic or the claim, chosen randomly, will be masked with a special [MASK] token and predicted by the language model. For the classification objective, the concatenated sequence without any modification is encoded by the language model, and the [CLS] token is used for classifying the pair. The classification objective is identical to the one used in a common clean finetuning setup for a classification task. In a noise-free scenario, using the classification objective may be enough for finetuning the language model. However, the conditional masked language modeling objective is able to allow the model to learn how a topic and a claim interacts conditioned on a noisy relation without forgetting how to do language modeling. Preliminary experiments show that this objective is very important in ensuring model performance. Shown in Section \ref{sec:data1_extraction}, the $D_1$ dataset is imbalanced with a large number of data points labeled as Neutral or Support. Random sampling with small weights on Support and Neutral is performed on this dataset such that there are 0.7 million data points for all classes, ensuring balanced training of all labels.

\subsection{Noisy and clean finetuning}
\label{sec:clean_finetuning}
After distant finetuning, the encoder from the pretrained language model is ready for a finetuning stage where training data is less noisy and more similar to data used in the downstream task. At this stage, the refined noisy dataset described in Section \ref{sec:data2_extraction}, combined with the original gold dataset and a dataset with gold data points back-translated from English, is used for training. Only the classification objective is used in this stage, resembling the common finetuning process. After two epochs, the encoder is ready for clean finetuning with the gold training set. In order to increase robustness of the model and regularize learning, a small portion of $D_2$ equal to 8\% of the gold training set is added into the gold training set for the final clean finetuning.

\subsection{Ensembling}

Due to the small size of gold training data, different random seeds yield models with varying performances. Randomness caused by the noisy data sampling process also causes models to be trained with different training sets thus having different performances. We propose to ensemble best-performing models trained with different configurations together, which leads to a final composite model with high robustness. The final prediction probabilities are calculated as the product of the prediction probabilities from all the models:
\begin{equation}
    p_{\text{final}}(\mathbf{y}|\mathbf{x}_i) = \prod_{j} p_{j}(\mathbf{y}|\mathbf{x}_i)
\end{equation}
where $i$ is the index of an input $\mathbf{x}$, and $\mathbf{y}$ is the output probabilities and $j$ is the index of a model in the ensemble.

\section{Experiments}

The datasets provided in the shared task include a training set with 6,416 data points and a development set with 990 data points, which are used for model development and hyperparameter tuning. For hyperparameters, we use the XLM RoBERTa large model \cite{Conneau2020-kz} as the base pretrained language model encoder in our classifier, which has 24 hidden layers, 16 attention heads with 4096 as intermediate embedding size and 1024 as the size of the final hidden embeddings. Dropout for all layers is set to be 0.1.

The classifier is first trained with the distant finetuning setup with $D_1$ and $D_x$ datasets for 58,500 steps with a batch size of 8 per step. A gradient update is performed every 4 steps, making the effective batch size to be 32. The learning rate for this stage is set to be $8 \times 10^{-6}$. The mix ratio between $D_1$ and $D_x$ is 4:1, meaning that 80\% of the time, a batch is sampled from $D_1$. After a batch is sampled from a dataset, a training objective is chosen randomly between classification and language modeling. 
For the noisy and clean finetuning, the number of epochs is chosen to be 2. The learning rate is $6 \times 10^{-6}$ and the batch size is 32.

AdamW \cite{Loshchilov2017-ms} is used as the optimizer at all stages. The top classification layer is re-initialized between stages. Performances of different experiment setups are reported in accuracy on the development set, because the test set is not released.

\subsection{Encoders}

We first examine performances of different pretrained languages models as the base encoder in clean finetuning. The goal of this experiment is to measure model performances when finetuned with gold data only. Table \ref{tab:encoders} shows the results of finetuning with a number of popular Chinese pretrained models \cite{Cui2020-pg} as well as the XLM-RoBERTa model. Interestingly, the only model that is not trained entirely on Chinese data, XLM-RoBERTa large, is the best performing model of all. This indicates that multilingual training is helpful even when the downstream task is in a specific language only. The Electra model, which has been reported to reach state-of-the-art performances on many language understanding tasks, is not able to outperform both RoBERTa large and XLM-RoBERTa. Finally, there is a substantial performance gap between smaller BERT base models and larger XLM-RoBERTa models, showcasing the importance of training data size for pretraining as well as objectives used in pretraining.

\begin{table}
\caption{Performance of different encoders with finetuning on the development set}\label{tab:encoders}
\begin{tabular}{|l|l|}
\hline
Encoder Type & Development accuracy \\
\hline
Chinese BERT wwm base & 76.22 \\
Chinese BERT wwm ext & 78.78  \\
Chinese Electra 180g large & 80.80 \\
Chinese RoBERTa wwm ext large & 82.61 \\
\bf XLM-RoBERTa large & \bf 85.24 \\
\hline
\end{tabular}
\end{table}

\subsection{Distance finetuning}

With XLM-RoBERTa large chosen as the encoder of the model, we explore the number of steps needed for the best performance with distant finetuning. Table \ref{tab:distant} shows the model performance on the development set with only distant finetuning with no gold training set used at all. Model performance increases steadily as the number of steps increases. A pretrained encoder with a randomly initialized classification layer gets 32.72 accuracy, but when distant finetuning is used, the model is able to reach 70.49, which is close to how Chinese BERT base performs with finetuning. This shows that the training signal in the dataset used in distant finetuning is very strong, and the model is able to learn robustly to detect stances of sentences, despite the fact that it has not seen any gold training data and there exists a style difference between the noisy dataset from the internet and human-authored gold training data. Finally, the model trained with 58500 steps is used for clean finetuning because of time constraints in the shared task, but it is likely that further improvement may be acquired with even more training steps.

\begin{table}
\caption{Performance of the model in distant finetuning on the development set with XLM-RoBERTa large}\label{tab:distant}
\begin{tabular}{|l|l|}
\hline
Number of distant finetuning steps & Development accuracy \\
\hline
0 & 32.72\\
16500 & 68.28\\
38500 & 70.20\\
\textbf{58500} & \textbf{70.49} \\
\hline
\end{tabular}
\end{table}

\subsection{Stages of finetuning}

Good performance from distant finetuning can be further improved by finetuning the model with gold training data. As described in Section \ref{sec:clean_finetuning}, two finetuning stages follow the distant finetuning, which both involve gold training data. Table \ref{tab:clean_finetune} shows how different combinations of finetuning stages affect model performance. The 3-stage finetuning is most effective in improving model performance and robustness, as it further increases model accuracy by 1.52 points compared to directly using clean finetuning after distant finetuning. Although a large amount of automatically generated data is used in noisy finetuning, model performance is only slightly lower than clean finetuning, showing both the high quality of the noisy data and high robustness of the model.

\begin{table}
\caption{Model performance with different combinations of finetuning stages.}\label{tab:clean_finetune}
\begin{tabular}{|l|l|}
\hline
Stage & Development accuracy \\
\hline
Distant finetuning & 70.49 \\
Distant + Noisy finetuning & 89.80 \\
Distant + Clean finetuning & 90.20 \\
\bf Distant + Noisy + Clean finetuning & \bf 91.72 \\
\hline
\end{tabular}
\end{table}

\subsection{Added noisy samples in finetuning}

We also look at if adding noisy samples into the clean training set in clean finetuning is able to help the model improve its performance, most likely by regularizing model training. Different numbers of noisy training data points from $D_2$ are randomly sampled and added to the gold training set, as shown in Table \ref{tab:noisy_in_clean}. Model performance averaged across 50 seeds is reported here. Using no noisy data in final clean finetuning yields lowest performance in general, and adding a small amount of noisy data does help model performance. Comparing to the whole gold training set with more than 6000 training instances, adding 500 noisy data points does not introduce too much noise but the regularizing effect from the noisy data points helps the model to be more robust to test items not found in training. 

\begin{table}
\caption{Model performance with different number of noisy data points added into the gold training set in the final clean finetuning. Performance numbers are average accuracy over 50 random seeds.}\label{tab:noisy_in_clean}
\begin{tabular}{|l|l|}
\hline
Number of samples from $D_2$ & Development accuracy \\
\hline
0 & 89.25 \\
250 & 89.49 \\
\bf 500 & \bf 89.51 \\
1000 & 89.35 \\
\hline
\end{tabular}
\end{table}

\end{CJK*}

\section{Conclusion}

A new method to extract data with silver labels from raw text to finetune a system for stance classification has been proposed in this paper . The reliance on specific discourse relations in the data extraction has ensured that the extracted silver topic and claim pairs are of high quality and the relations between the extracted pairs are relevant to the stance classification task. In order to use such silver data, a 3-stage training scheme where the noisy level in the training data decreases over different stages going from most noisy to least noisy is also proposed in the paper. We show through detailed experiments that the automatically annotated dataset as well as the 3-stage training help improve model performance in stance classification. Our approach ranks 1$^{\text{st}}$ among 26 competing teams in the stance classification track of the NLPCC 2021 shared task Argumentative Text Understanding for AI Debater, which confirms the effectiveness of our approach.

%
%
%
\bibliographystyle{splncs04}
\bibliography{stance}
\end{document}